\documentclass[letterpaper]{article} % DO NOT CHANGE THIS
\usepackage{aaai23}  % DO NOT CHANGE THIS
\usepackage{times}  % DO NOT CHANGE THIS
\usepackage{helvet}  % DO NOT CHANGE THIS
\usepackage{courier}  % DO NOT CHANGE THIS
\usepackage[hyphens]{url}  % DO NOT CHANGE THIS
\usepackage{graphicx} % DO NOT CHANGE THIS
\usepackage{natbib}  % DO NOT CHANGE THIS AND DO NOT ADD ANY OPTIONS TO IT
\usepackage{caption} % DO NOT CHANGE THIS AND DO NOT ADD ANY OPTIONS TO IT
\usepackage{algorithm}
\usepackage{algorithmic}

\usepackage{newfloat}
\usepackage{listings}
\DeclareCaptionStyle{ruled}{labelfont=normalfont,labelsep=colon,strut=off} % DO NOT CHANGE THIS
\floatstyle{ruled}
\newfloat{listing}{tb}{lst}{}
\floatname{listing}{Listing}

\usepackage{tikz}
\usepackage{comment}
\usepackage{amsmath,amssymb} % define this before the line numbering.
\usepackage{booktabs}
\usepackage{multirow}
\usepackage{colortbl}
\usepackage{caption}
\usepackage{pifont}% 
\newcommand{\cmark}{\ding{51}}%
\title{Geometry-Aware Network for Domain Adaptive Semantic Segmentation}
\author{
    Yinghong Liao\textsuperscript{\rm 1,\rm 2}\equalcontrib,
    Wending Zhou\textsuperscript{\rm 1,\rm 2}\equalcontrib,
    Xu Yan\textsuperscript{\rm 1,\rm 2},
    Shuguang Cui\textsuperscript{\rm 2,\rm 1},
    Yizhou Yu\textsuperscript{\rm 3},
    Zhen Li\textsuperscript{\rm 2,\rm 1}\thanks{Corresponding author.}
}
\affiliations{
    \textsuperscript{\rm 1} The Future Network of Intelligence Institute, The Chinese University of Hong Kong (Shenzhen)\\
    \textsuperscript{\rm 2} School of Science and Engineering, The Chinese University of Hong Kong (Shenzhen)\\
    \textsuperscript{\rm 3} Department of Computer Science, The University of Hong Kong\\
    \{yinghongliao@link., lizhen@\}cuhk.edu.cn
}

\usepackage{bibentry}
\begin{document}

\maketitle

\begin{abstract}
Measuring and alleviating the discrepancies between the synthetic (source) and real scene (target) data is the core issue for domain adaptive semantic segmentation.
Though recent works have introduced depth information in the source domain to reinforce the geometric and semantic knowledge transfer, they cannot extract the intrinsic 3D information of objects, including positions and shapes, merely based on 2D estimated depth.
In this work, we propose a novel {\textbf{G}}eometry-{\textbf{A}}ware {\textbf{N}}etwork for {\textbf{D}}omain {\textbf{A}}daptation ({\textbf{GANDA}}), leveraging more compact 3D geometric point cloud representations to shrink the domain gaps.
In particular, we first utilize the auxiliary depth supervision from the source domain to obtain the depth prediction in the target domain to accomplish structure-texture disentanglement.
Beyond depth estimation, we explicitly exploit 3D topology on the point clouds generated from RGB-D images for further coordinate-color disentanglement and pseudo-labels refinement in the target domain.
Moreover, to improve the 2D classifier in the target domain, we perform domain-invariant geometric adaptation from source to target and unify the 2D semantic and 3D geometric segmentation results in two domains.
Note that our GANDA is plug-and-play in any existing UDA framework.
Qualitative and quantitative results demonstrate that our model outperforms state-of-the-arts on GTA5$\to$Cityscapes and SYNTHIA$\to$Cityscapes.
\end{abstract}

\section{Introduction}

Semantic segmentation has obtained admirable progress owing to the success of deep learning in computer vision.
However, the pixel-level annotations on large-scale image data require time-consuming and laborious manual work.
To overcome this challenge, the synthetic datasets whose photo-realistic images and annotations are automatically generated via computer graphics,~\emph{e.g.}, GTA5~\cite{StephanR2016GTA} and SYNTHIA~\cite{Ros2016SYN}, are leveraged to explore training on the synthetic images instead of real-scene ones (synthetic-to-real).
Nevertheless, the model trained on synthetic data exhibits poor generalization on real-world images, \emph{i.e.}, the domain discrepancy occurs.
The discrepancy lies in the differences between the synthetic and real-scene data, \emph{e.g.}, the contrast, lighting, object shape, and surface textures.
Therefore, an effective Unsupervised Domain Adaptation (UDA) model, where the target annotations are not provided, is needed.

\begin{figure}[t]
    \centering\includegraphics[width=0.98\columnwidth]{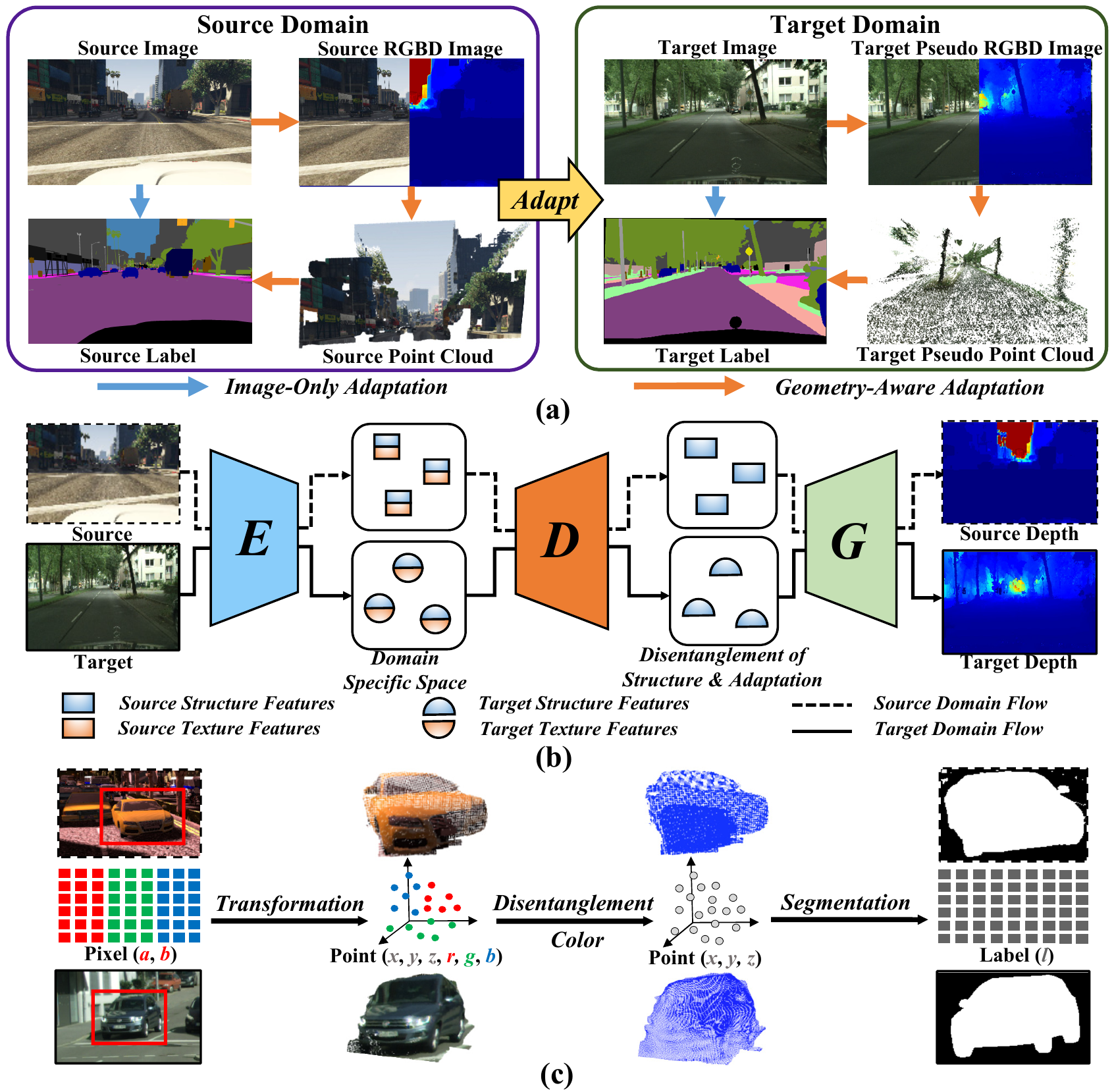}
        \caption{(a) Illustrations of the traditional image-only adaptation and our proposed geometry-aware adaptation. Geometry-aware adaptation only utilizes the depths in the source domain and leverages the domain-invariant geometric information in the point cloud transformed from RGB-D images. (b) The disentanglement of structural features in the depth-aware domain adaptation. (c) The disentanglement of spatial coordinates and RGB colors in the 3D points transformed from 2D pixels.}
        \label{fig:fig1}
\end{figure}

Instead of aligning the domain distributions that may generate sub-optimal solutions~\cite{Yang2020APODA,Vu2018ADVENT}, the recent UDA works~\cite{Zhang2021ProDA,Hoyer2022DAFormer,Hoyer2022HRDA} adopt self-training to exploit the target information: first predicting pseudo labels based on the trained parameters from the annotated data and then rectifying the pseudo labels via the target knowledge.
%
% Specifically, ProDA~\cite{Zhang2021ProDA} calculates the class-wise feature centroids, \emph{i.e.}, prototypes, on the pseudo labels to reweight the predictions and utilizes self-distillation to improve the performance by a large margin.
% %
Besides, depth-based multi-task learning~\cite{Vu2019DADA,Wang2021CorDA} also has demonstrated the advantages in utilizing the target image features.
% %
% CorDA~\cite{Wang2021CorDA} incorporates the auxiliary task of depth estimation to ease the gap between synthetic and real data, then utilizes the depth prediction differences between source and target images to perform reweighting on the target pseudo maps.
%
Although the self-training and depth-based methods have achieved significant progress in domain adaptive semantic segmentation, they still have several issues:
\textbf{1)} The predicted low-confidence pseudo labels have uncertainty in the patches of the object edges and might lead to assigning incorrect labels to these edge pixels, especially for the small objects.
\textbf{2)} Despite the effectiveness in extracting structure knowledge, auxiliary depth estimation cannot accurately manifest the object spatial positions and the inherent 3D geometric object shapes, leading to the unsatisfying performances in the small objects.
\textbf{3)} The depth-based methods~\cite{Wang2021CorDA} obtain the depths for the target Cityscapes images via the self-supervised depth estimation and utilizes them as the ground truth to supervise the multi-task training.
However, when the target dataset consists of only the images, the target depths are not available to perform a similar self-supervised depth estimation.

To address the above issues, in this work, we propose to incorporate geometric information into the prevalent image-only UDA frameworks and learn a complete 3D geometric object shape with the only source depth information, as shown in Figure~\ref{fig:fig1}(a).
Since the depth information is accessible in the generation of synthetic images, we consider leveraging the source depths to estimate the target depths.
As displayed in Figure~\ref{fig:fig1}(b), the generation of depths can be seen as disentanglement of the object structures from the RGB images without colors and textures~\cite{Chang2019structure}.
Thus, the depth information from the source and target domains shares more domain-invariant geometric structure knowledge and avoids influences from the domain-specific textures.
Furthermore, with the 2D CNNs, the mere depth estimation is insufficient to exploit the implicit 3D geometric shape knowledge in RGB and Depth (RGB-D) images.
To facilitate the adaptation of the geometric information, we convert the RGB-D images to more compact 3D scenes in point clouds.
With point clouds and 3D point-based networks, the 3D geometric topology on the source and target domains can be exploited, including the explicit object positions and distinctive object shapes.
As shown in Figure~\ref{fig:fig1}(c), the 3D coordinates and colors can be disentangled, and the 3D spatial coordinates that represent the domain-invariant object shapes can be explored independently.

Accordingly, we propose a novel plug-and-play {\textbf{G}}eometry-{\textbf{A}}ware {\textbf{N}}etwork for {\textbf{D}}omain {\textbf{A}}daptation ({\textbf{GANDA}}) for the UDA segmentation task.
GANDA consists of two stages: depth-aware adversarial adaptation and geometry-aware adaptation.
In the depth-aware adversarial adaptation stage, GANDA first accomplishes the depth estimation on the annotated source images, then predicts the target depths based on the learned parameters.
With the obtained target pseudo depths, the target RGB-D images are generated, and the corresponding point clouds are reconstructed.
In the geometry-aware adaptation stage, GANDA leverages 3D point clouds to perform 2D semantic and 3D geometric segmentation simultaneously.
Specifically, we design a Geometric Adaptation Module to align the domain-invariant geometric information on 3D point clouds from source to target.
The target predicted labels are enhanced with the geometric information by unifying the semantic and geometric segmentation results on two domains.
Geometry-aware adaptation is especially effective in segmentation of the small-object classes.
The main contributions of this paper are as follows:
\begin{itemize}
    \item We propose GANDA, a novel plug-and-play UDA segmentation framework to leverage the complete 3D geometric representations with only source depth image.
    This model introduces domain-invariant geometric information based on the structure-texture and coordinate-color disentanglements.
    \item The Geometry-Aware Adaptation stage is introduced to align the domain-invariant geometric knowledge from source to target on the reconstructed point clouds from RGB-D images.
    \item The proposed GANDA outperforms the state-of-the-art methods on the classic synthetic-to-real segmentation tasks, including GTA5 $\to$ Cityscapes and SYNTHIA $\to$ Cityscapes.
\end{itemize}
\begin{figure*}[th]
  \centering
  \includegraphics[width=2.00\columnwidth]{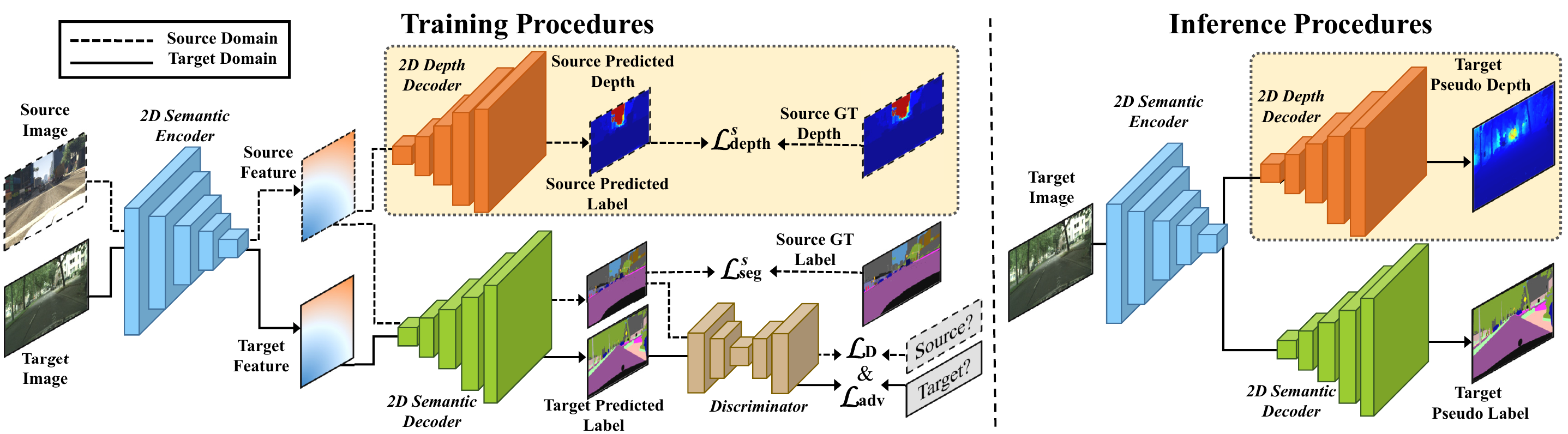}
  \caption{Depth-Aware Adversarial Adaptation. The proposed plug-and-play depth-aware adaptation modules are highlighted in rounded rectangles with dotted lines.}
  \label{fig:fig2}
\end{figure*}
\section{Related Work}

\noindent\textbf{Unsupervised Domain Adaptation in Segmentation.}
Appropriate measurement of the domain gap and accurate disentanglement of the domain-invariant knowledge are the keys to effective Unsupervised Domain Adaptation (UDA) in the synthetic-to-real semantic segmentation task.
Accordingly, two mainstream approaches come to stage: domain alignment and knowledge extraction from the target domain.
Based on adversarial learning, the first commonly used methods perform domain alignment to find domain-invariant features from different levels, including image level~\cite{Chen2019CrDoCo,Ma2021}, feature level~\cite{Nuriel2021}, and semantic level~\cite{Zhang2021ProDA}.
Though effective in narrowing the domain gap, these works fail to achieve better segmentation results since they focus on the information from source domain without further mining the knowledge from target domain.
By comparison, the latter ones~\cite{Yang2020LDR,Zhang2021ProDA,Hoyer2022HRDA} seizes the domain-specific knowledge from target domain to enhance the pseudo labels.

\noindent\textbf{Depth-based Multi-task Learning in UDA.}
Some researchers~\cite{Lee2018spigan,Vu2019DADA} resort to the auxiliary tasks to reinforce knowledge mining and transfer,~\emph{e.g.}, depth estimation.
%
% SPIGAN~\cite{Lee2018spigan} applies the GAN~\cite{Goodfellow2014GAN} to perform style transfer from source to target, in which a depth regression task is adopted to regularize the generator.
%
DADA~\cite{Vu2019DADA} introduces an auxiliary depth regression task to obtain the dense depth knowledge from the source domain to facilitate the adaptation.
GIO-Ada~\cite{Chen2019} proposes to utilize the auxiliary geometric information to ease the domain gap on both the input and output level.
However, these methods are based on adversarial learning and their performances are limited.
Thus the advantages of appending auxiliary information or tasks are not prominent.
Very recently, CorDA~\cite{Wang2021CorDA} uses the depth information on both domains to supervise the training of two depth decoders.
Though CorDA outperforms the previous works that employ depth estimation significantly, it allows the supervision from target depth maps that might be inaccessible.

\noindent\textbf{3D Point Cloud Segmentation.}
The methods of 3D point cloud segmentation are primarily divided into two categories: point-based methods~\cite{Qi2017PointNet,Qi2017PointNet++,Thomas2019kpconv,yan2020pointasnl} and voxel-based methods~\cite{Wu2015ShapeNet,Choy2019Minkowski,yan2021sparse,Tang2020Searching,yan20222dpass,Zhu2021Cylindrical}.
In the point-based methods, PointNet~\cite{Qi2017PointNet} and its improved PointNet++~\cite{Qi2017PointNet++} are the pioneers of applying permutation-invariant set functions on the raw 3D geometric points based on the MLP networks.
%
% KPConv~\cite{Thomas2019kpconv} performs the kernel point convolution on the point clouds, but it has the problem of time-consuming inference on the scenes containing large amounts of points.
%
Early voxel-based methods~\cite{Wu2015ShapeNet,Choy2019Minkowski} convert the points to voxels and utilize 3D convolutions to process the voxels.
Recently, Cylinder3D~\cite{Zhu2021Cylindrical} proposes an asymmetrical residual block to reduce computational costs of 3D convolutions.
Since the point clouds generated by RGB-D images are single-view and have more irregular shapes than the usual point clouds, our model applies the point-based model to extract 3D geometric information from point clouds.

\section{Our Approach}
In this section, we propose GANDA for domain adaptive semantic segmentation, which applies a two-stage self-training pipeline, including depth-aware adversarial adaptation and geometry-aware adaptation.
\subsection{Preliminaries}
\label{sec:sec3.1}
In the UDA segmentation task, we denote the source dataset containing $n^{s}$ images as $\mathbf{X}^{s} = \{\mathbf{x}^{s} \in \mathbb{R}^{3 \times H \times W}\}^{n^{s}}_{i=1}$ and the corresponding semantic labels as $\mathbf{Y}^{s} = \{\mathbf{y}^s \in \{0, 1\}^{n_c \times H \times W}\}^{n^{s}}_{i=1}$, where each image $\mathbf{x}^s $ and label $\mathbf{y}^s$ share the same height of $H$ and the width of $W$, and $n_c$ represents the number of classes.
Accordingly, the target dataset has $n^{t}$ images, $\mathbf{X}^{t} = \{\mathbf{x}^{t}\}^{n^{t}}_{j=1}$.
Since the only supervision from the source domain is limited, self-training~\cite{Pan2020self} is exploited to handle the unlabeled dataset in the semi-supervised learning and UDA tasks.
After training multiple steps on the source domain, the model is able to produce the coarse pseudo labels on the target domain, $\mathbf{\hat{Y}}^t = \{\mathbf{\hat{y}}^t\}^{n^{t}}_{j=1} = \{\mathop{\arg \max} \mathbf{p}^t\}^{n_{t}}_{j=1}$, where $\mathbf{p}^t$ is the softmax probability of classes for the target image $\mathbf{x}^t$.
The target pseudo label $\mathbf{\hat{y}}^t$ for $\mathbf{x}^{t}$ is obtained from the class that has the highest soft confidence scores.
With the source label $\mathbf{y}^s$ and the target pseudo label $\mathbf{\hat{y}}^t$, the Cross-Entropy (CE) loss in self-training segmentation is calculated as:
\begin{equation}
    \begin{aligned}
    \mathcal{L}_{\text{seg}} = -\sum_{i=1}^{HW} \sum_{k=1}^{n_c} \mathbf{y}^s_{ik} \log \mathbf{p}^s_{ik}  - \sum_{j=1}^{HW}\sum_{k=1}^{n_c} \mathbf{\hat{y}}^t_{jk} \log \mathbf{p}^t_{jk}.
    \end{aligned}
    \label{eq:eq1}
\end{equation}
% The generated pseudo labels $\mathbf{\hat{Y}}^t$ are noisy and uncertain.
% %
% The uncertainty in the pseudo labels might affect the final segmentation results, so many works~\cite{Zheng2021Uncertainty,Zhang2021ProDA} try to rectify the pseudo labels via exerting more optimizing constraints on the target domain.

\subsection{Depth-Aware Adversarial Adaptation}
\label{sec:sec3.2}
\begin{figure*}[t]
  \centering
  \includegraphics[width=1.99\columnwidth]{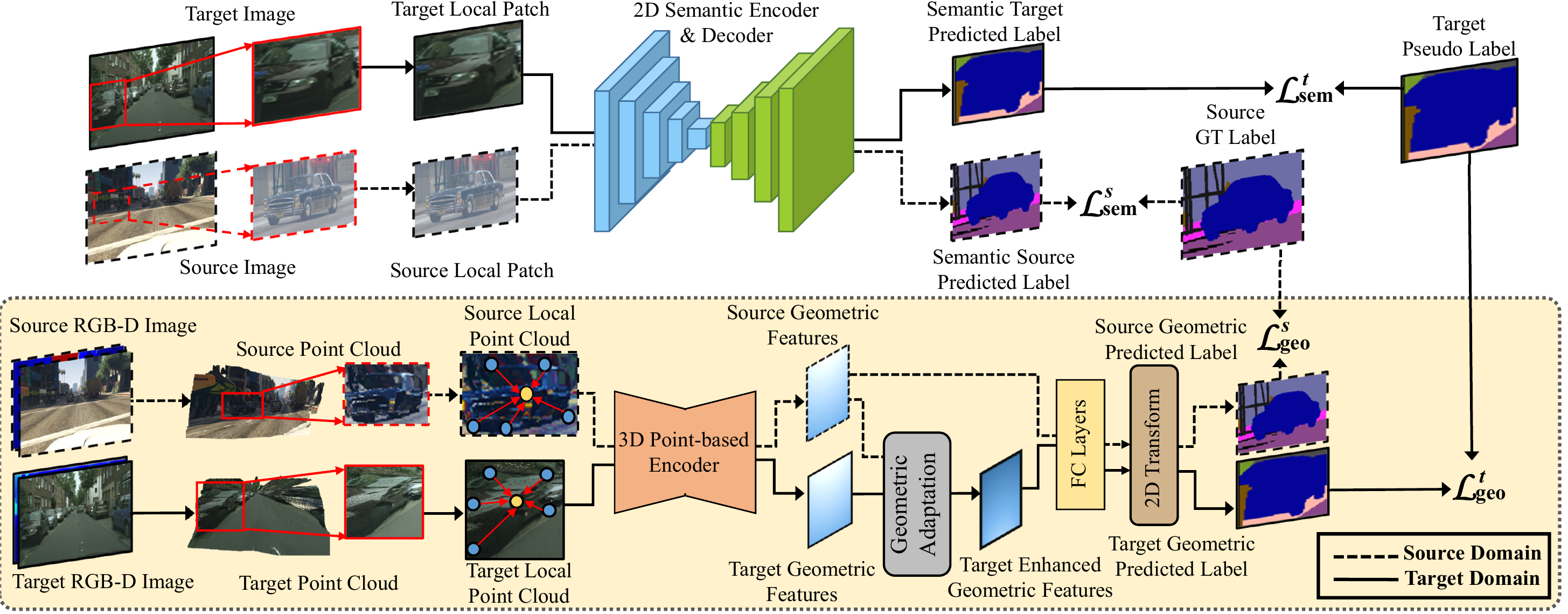}
  \caption{Geometry-Aware Adaptive Training Scheme. The proposed plug-and-play geometry-Aware adaptation pipeline is highlighted in the dotted rounded rectangle.}
  \label{fig:fig3}
\end{figure*}
In this work, other than the image-only adaptation, we take the accessible depth maps in the source domain into account.
Specifically, as shown in Fig~\ref{fig:fig1}(b), our GANDA first learns to disentangle the structural information from single source RGB images $\mathbf{X}^{s}$ under the supervision of source depths $\mathbf{D}^{s} = \{\mathbf{d}^s \in \mathbb{R}^{H \times W}\}^{n^{s}}_{i=1}$.
Then it applies the learned depth knowledge to generate pseudo depth maps $\mathbf{\hat{D}}^{t} = \{\mathbf{\hat{d}}^t \in \mathbb{R}^{H \times W}\}^{n^{t}}_{j=1}$ on the raw target images $\mathbf{X}^{t}$.
%
% Though depth estimation is an ill-posed problem and the predicted depth for each pixel might be inaccurate, the significant depth disparities in the object edges and the small depth gradients within objects ensure that the object's geometric shapes are maintained in the target depth maps.
% %
% As displayed in Fig~\ref{fig:fig1}(c), the predicted depth for the object ``car" may vary in different images, but the disparity of the depths ensures the shape of ``car" is complete and recognizable after 3D transformation.
%
In the training part of the depth-aware adversarial adaptation, as displayed in Figure~\ref{fig:fig2}, we incorporate the depth-aware knowledge transfer into the classic adversarial UDA.
%
% The purposes of this stage are:
% %
% (1) to obtain the basic model for recognizing the objects in the scenes;
% %
% (2) to produce more reliable pseudo labels on the target domain with the trained model;
% %
% (3) to generate accurate pseudo depth maps on the target domain for the following geometry-aware adaptation. 

The depth-aware adversarial adaptation adopts adversarial learning that 2D semantic encoder is optimized to ``fool" the discriminator with the help of source depth estimation:
\begin{equation}
\small
    \begin{aligned}
        \mathcal{L}_{\text{D}} = -\sum_{i=1}^{HW}\!\sum_{k=1}^{n_c}\!\mathbf{b}^s_{ik} \log &P(\mathbf{D}_{ik}\!=\!0)\!-\sum_{j=1}^{HW}\!\sum_{k=1}^{n_c}\!\mathbf{b}^t_{jk}\!\log\!P(\mathbf{D}_{jk}\!=\!1),
        \end{aligned}
    \label{eq:eq2}
\end{equation}
\begin{equation}
    \begin{aligned}
        \mathcal{L}_{\text{adv}} &= -\sum_{j=1}^{HW} \sum_{k=1}^{n_c} \mathbf{b}^t_{jk} \log P(\mathbf{D}_{jk} = 0)
    \end{aligned}
    \label{eq:eq3}
\end{equation}
where $\mathbf{b}^s_k$ and $\mathbf{b}^t_k$ are the binary indicator (source or target) of class $k$, $P(\mathbf{D}_{k} = 0)$ and $P(\mathbf{D}_{k} = 1)$ represent the predicted probability of class $k$ from the source or target domain, respectively.

% Equipped with accurate predictions from the segmentation parts, depth-aware adversarial adaptation simultaneously performs depth estimation on the source domain.
% %
The 2D depth decoder is trained simultaneously using the depth loss $\mathcal{L}_{\text{depth}}^{s}$ between the predicted depth map $\mathbf{\widetilde{d}}^{s}$ and the source depth map $\mathbf{{d}}^{s}$:
\begin{equation}
    \begin{aligned}
        \mathcal{L}_{\text{depth}}^{s} &= \mathcal{L}_{\text{SILog}}^{s} + \lambda \mathcal{L}_{\text{grad}}^{s},
    \end{aligned}
    \label{eq:eq4}
\end{equation}
where $\lambda$ is the loss weight.
$\mathcal{L}^{s}_{\text{SILog}}$ is the Scale-Invariant Logarithmic loss for depth estimation~\cite{Yuan2022NeWCRFs}:
\begin{equation}
    \begin{aligned}
        \Delta \mathbf{d}^{s}_{i} &= \log \mathbf{\widetilde{d}}^{s}_{i}- \log \mathbf{d}^{s}_{i},\\
        \mathcal{L}^{s}_{\text{SILog}} &= 
        \psi\sqrt{\frac{1}{K}\sum_{i=1}^{HW}\Delta \mathbf{d}^{s~2}_{i} - \frac{\gamma}{K}(\sum_{i=1}^{HW}\Delta \mathbf{d}^{s}_{i})^2},
    \end{aligned}
    \label{eq:eq5}
\end{equation}
where $\psi$ is the scale
constant set as $10$ and $\gamma$ denotes the variance minimizing factor set as $0.85$.
$\mathcal{L}_{\text{grad}}$ is the smoothness loss with the image gradients from two directions, $g_x$ and $g_y$.
We train the whole stage using the following total loss with the weight $\mu$:
\begin{equation}
    \begin{aligned}
        \mathcal{L}_{\text{depth-aware}} = \mathcal{L}^{s}_{\text{seg}} + \mathcal{L}^{s}_{\text{depth}} +
        \mu \mathcal{L}_{\text{adv}}.
    \end{aligned}
    \label{eq:eq7}
\end{equation}
In the inference part in Figure~\ref{fig:fig2}, the target pseudo label $\mathbf{\hat{y}}^t$ and pseudo depth map $\mathbf{\hat{d}}^t$ are generated for the following self-training based geometry-aware adaptation.

\subsection{Geometry-Aware Adaptation}
\label{sec:sec3.3}

% \noindent\textbf{Why are 2D CNNs Inefficient in Alleviating Domain Gaps?}
% %
% 2D CNNs derive the features via the convolutions on the local pixels and focus on semantic information.
% %
% However, since the images are the projection of the 3D natural scenes and two close pixels might have large distances in 3D space, 2D CNNs are deficient in restoring the original 3D geometry.
% %
% Although the depth-aware adversarial adaptation enables the disentanglement between the structures and the textures, as shown in Fig~\ref{fig:fig1}(b), 2D CNNs still cannot unearth the inherent 3D object information, including the spatial object positions and geometric object shapes.
% %
% % Therefore, the domain discrepancies in the 3D topology also exist, and the 2D classifier cannot capture more spatial features.
% %
% Therefore, to construct a more compact geometric knowledge that facilitates the segmentation, we resort to 3D spatial point clouds.
% \newline
% \newline
\noindent\textbf{From 2D RGB Pixels to 3D Geometric Points.}
Given the source map $\mathbf{d}^{s}$ and the target pseudo depth map $\mathbf{\hat{d}}^{t}$, we transform a depth map $\mathbf{d}$ and the RGB image $\mathbf{X}$ to an RGB-D image and further produce the corresponding point cloud $\mathbf{V} = \{\mathbf{v} \in \mathbb{R}^{6}\}_{k=1}^{n_v}$ for each image, where $n_v$ represents the number of the points $\mathbf{v}$.
According to the mapping relationship between the RGB pixels and the 3D points, $n_v = HW$, and each point corresponds to one pixel in the original image $\mathbf{X}$.
In 3D point cloud processing, a point $\mathbf{v}$ consists of the 3D spatial coordinate $(x, y, z)$ and the RGB color information$(r, g, b)$ as illustrated in Figure~\ref{fig:fig1}(c), $\mathbf{V} = \{ \mathbf{v} \in \mathbb{R}^{6}\}_{k=1}^{n_v} = \{(x, y, z, r, g, b) \in \mathbb{R}^{6}\}_{k=1}^{n_v}$.
%
% We further present the transformed point clouds of ``cars" from different domains as examples in Fig~\ref{fig:fig1}(c), where the shapes of the cars are preserved after removing colors.
%
The geometric coordinate $(x, y, z)$ is seen as the domain-invariant information of the object and disentangled from $\mathbf{v}$.

\noindent\textbf{Disentanglement of the 3D Geometric Coordinates and RGB Colors.} In the geometry-aware adaptation, as displayed in Figure~\ref{fig:fig3}, we discard the trained 2D depth decoder in the previous stage and integrate a 3D point-based encoder that performs disentanglement on the spatial geometric features.
%
% Unlike 2D feature extractor that emphasizes semantics, the 3D point-based encoder $\mathbf{\Phi}$ possesses more 3D geometric information.
%
% The PointNet++~\cite{Qi2017PointNet++} is adopted to implement the 3D point-based encoder $\mathbf{\Phi}$.
%
Initially, the input point $\mathbf{v}$ is decomposed into the geometric coordinates $\mathbf{h} = (x, y, z)$ and the point feature $\mathbf{v} = (x, y, z, r, g, b)$, which isolates the spatial positional information individually and also preserves the RGB color information.
%
% The 3D point-based encoder first applies the iterative farthest point sampling (FPS) to sample several small point sets, and select the most distant point from others within the sampled points as the centroids $\mathbf{Z}$.
% %
% With the sampled point centroids $\mathbf{Z}$, the point set $\mathbf{V}$ is grouped locally via k-Nearest Neighbor (KNN) algorithms, and groups of point set $\mathbf{G}$ with $K$ neighbor points are generated for these centroids.
%
The 3D point-based encoder $\mathbf{\Phi}$ takes the local regions of points as the inputs, including geometric coordinates $\mathbf{H} =\{\mathbf{h}\}_{k=1}^{n_v}$ and point features $\mathbf{V} = \{\mathbf{v}\}_{k=1}^{n_v}$, and obtains geometry-aware features:
\begin{equation}
    \begin{aligned}
        \mathbf{F}_{\text{geo}} = \mathbf{\Phi}(\mathbf{H}, \mathbf{V}) = \{\mathbf{\Phi}(\mathbf{h}, \mathbf{v})\}_{k=1}^{n_v} = \{\mathbf{f}_{\text{geo}} \in \mathbb{R}^{m}\}_{k=1}^{n_v},
    \end{aligned}
    \label{eq:eq8}
\end{equation}
where $m$ is the size of geometric feature $\mathbf{f}_{\text{geo}}$.
Note that all the operations in the 3D point-based encoder are explicitly performed on the 3D spatial coordinates $\mathbf{H}$, which guarantees that the features $\mathbf{F}_{\text{geo}}$ maintain the critical 3D geometric information to the largest extent.

\begin{figure}[t]
  \centering
  \includegraphics[width=1.00\columnwidth]{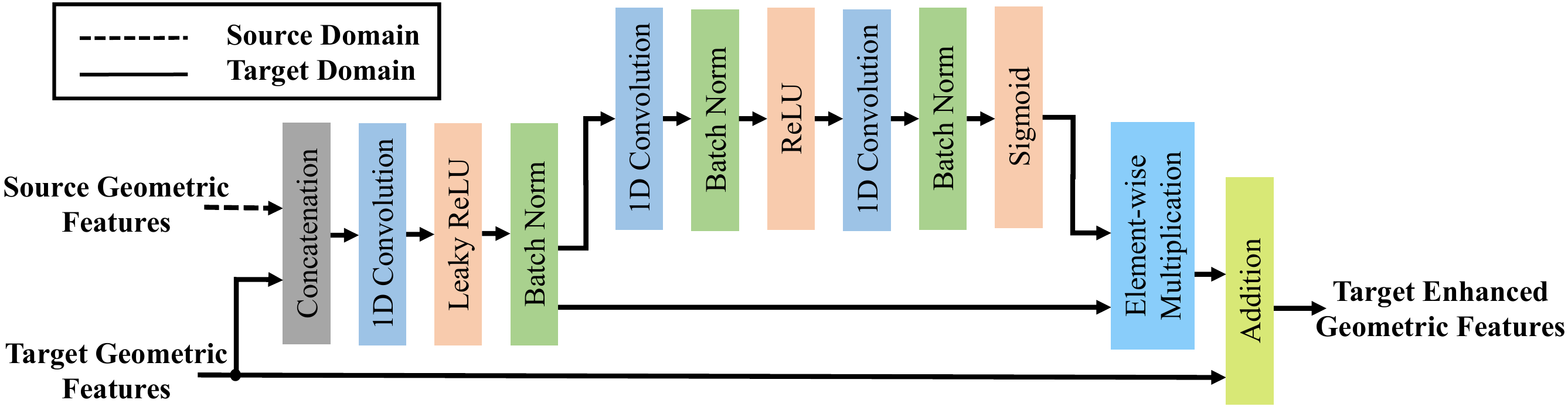}
  \caption{Geometry Adaption (GA) module. It computes the geometric shape similarity of the objects, and extracts the geometric features of the objects with similar shapes.}
  \label{fig:fig4}
\end{figure}

\noindent\textbf{Geometry-Aware Adaptive Training Scheme.}
Our proposed GANDA is plug-and-play, and the Geometry-Aware Adaptive Training (GAAT) scheme can be appended to the original 2D adaptive segmentation scheme in parallel.
As displayed in Figure~\ref{fig:fig3}, the GAAT performs adaptive segmentation on the cross-modal inputs.
%
% In consideration of the effectiveness and efficiency, the 3D point-based encoder trains on the local part of a scene instead of the whole scene.
%
In the 2D adaptive part, the input source and target images are randomly cropped to produce local patches $\mathbf{x}^s_{\text{local}}$ and $\mathbf{x}^t_{\text{local}}$ with a smaller size.
Accordingly, in the 3D adaptive part, the cropping operation is performed on the same patch of the RGB-D images, then the local point clouds $\mathbf{V}^s_{\text{local}}$ and $\mathbf{V}^t_{\text{local}}$ are obtained after the pixel-to-point transformation.
The local patches and point clouds $\mathbf{x}_{\text{local}}$ and $\mathbf{V}_{\text{local}}$ are then fed to the 2D semantic model (encoder and decoder) and 3D point-based encoder $\mathbf{\Phi}$, respectively.
The outputs of the 2D semantic model are semantic-aware labels $\mathbf{\widetilde{y}}^{s}_{\text{sem}}$ and $\mathbf{\widetilde{y}}^{t}_{\text{sem}}$.
The outputs of the 3D point-based encoder $\mathbf{\Phi}$ are geometry-aware features $\mathbf{F}^s_{\text{geo}} = \mathbf{\Phi}(\mathbf{H}^s, \mathbf{V}^s)$ and $\mathbf{F}^t_{\text{geo}} = \mathbf{\Phi}(\mathbf{H}^t, \mathbf{V}^t)$.
We further propose a Geometric Adaptation (GA) module in Figure~\ref{fig:fig4} to transfer geometric knowledge from $\mathbf{F}^{s}_{\text{geo}}$ to $\mathbf{F}^{t}_{\text{geo}}$ and obtain the enhanced target feature $\mathbf{\widetilde{F}}^{t}_{\text{geo}}$.
The geometry-aware features $\mathbf{F}^{s}_{\text{geo}}$ and $\mathbf{\widetilde{F}}^{t}_{\text{geo}}$ are processed by two FC Layers and then transformed to 2D pixel space as $\mathbf{\widetilde{y}}^{s}_{\text{geo}}$ and $\mathbf{\widetilde{y}}^{t}_{\text{geo}}$.

\noindent\textbf{Geometric Adaptation Module.}
After the structure-texture and coordinate-color disentanglements, the 3D domain-invariant geometric information is isolated and can be utilized to perform adaptation to alleviate the domain gaps.
Based on this motivation, we propose a Geometric Adaptation (GA) module $\mathbf{\Gamma}$ in Figure~\ref{fig:fig4} to transfer geometric knowledge from $\mathbf{F}^{s}_{\text{geo}}$ to $\mathbf{F}^{t}_{\text{geo}}$ and derive the enhanced target feature $\mathbf{\widetilde{F}}^{t}_{\text{geo}} = \mathbf{\Gamma}(\mathbf{F}^{s}_{\text{geo}}, \mathbf{F}^{t}_{\text{geo}})$.
Since the source and target patch may be different objects, this module is designed as a residual form with the attention operation.
The GA module first performs concatenation on the geometric features $\mathbf{F}^{s}_{\text{geo}}$ and $\mathbf{F}^{t}_{\text{geo}}$.
Then the concatenated features $[\mathbf{F}^{s}_{\text{geo}}, \mathbf{F}^{t}_{\text{geo}}]$ are further performed the attention operation.
The derived feature is finally added to the target geometric features $\mathbf{F}^{t}_{\text{geo}}$.
While being trained on the abundant image patches, the GA module computes the geometric shape similarity of objects, where the geometric features of the objects with similar shapes are effectively extracted by the attention operation.

\noindent\textbf{Loss Functions.}
The semantic and geometric softmax probabilities of classes, $\mathbf{p}^{s}_{\text{sem}}$, $\mathbf{p}^{t}_{\text{sem}}$, $\mathbf{p}^{s}_{\text{geo}}$, and $\mathbf{p}^{t}_{\text{geo}}$, are computed training losses with the source ground truth label $\mathbf{y}^{s}_{\text{local}}$ and target pseudo labels $\mathbf{\hat{y}}^{t}_{\text{local}}$, respectively.
To increase the contribution of the long-tail classes, we derive the weights for each category $c$ based on its corresponding pixel numbers.
We first compute the frequency $\mathbf{\omega}_{c}^s$ of the pixel that belongs to the class $c$ on the source domain, then we obtain the weights for each class and assign larger values to the long-tail classes $\mathbf{w}_{c}^s = (\frac{\sum^{n_c}_{k=1}\mathbf{\omega}_{k}^s}{\mathbf{\omega}_{c}^s})^{\frac{1}{3}}$, $\mathbf{\omega}_{c}^s = \frac{1}{n_{s}HW}\sum^{n_s}_{i=1}\sum^{HW}_{j=1} \mathbf{m}_{cij}^s$, where $\mathbf{m}_{cij}^s$ is the mask at the pixel $j$ that belongs to class $c$, $\mathbf{m}_{cij}^s = 1$.
We add the class weights $\mathbf{w}^s$ to Eq.~\ref{eq:eq1}, and further derive the semantic and geometric segmentation loss:

\begin{equation}
\small
\begin{aligned}
    \mathcal{L}^{s}_{\text{seg}}&= \mathcal{L}^{s}_{\text{sem}} + \mathcal{L}^{s}_{\text{geo}} \\ &= \!-\sum_{i=1}^{HW} \sum_{k=1}^{n_c} \mathbf{w}_k^s \mathbf{y}^s_{\text{local}} \log \mathbf{p}^s_{\text{sem}}\!-\sum_{i=1}^{HW} \sum_{k=1}^{n_c} \mathbf{w}_k^s \mathbf{y}^s_{\text{local}} \log \mathbf{p}^s_{\text{geo}},\\
    \mathcal{L}^{t}_{\text{seg}}&= \mathcal{L}^{t}_{\text{sem}} + \mathcal{L}^{t}_{\text{geo}} \\ &= \!-\sum_{j=1}^{HW} \sum_{k=1}^{n_c} \mathbf{\hat{y}}^t_{\text{local}} \log \mathbf{p}^t_{\text{sem}}\!-\sum_{j=1}^{HW} \sum_{k=1}^{n_c} \mathbf{\hat{y}}^t_{\text{local}} \log \mathbf{p}^t_{\text{geo}}.
    \end{aligned}
    \label{eq:eq11}
\end{equation}
Furthermore, to minimize the semantic and geometric perceptual differences on the source and target domains, we employ the Kullback–Leibler (KL) divergence to design the Semantic-Geometric Consistency (SGC) loss $\mathcal{L}_{\text{SGC}}$:
\begin{equation}
    \begin{aligned}
    \mathcal{L}_{\text{SGC}}^{s} &= \sum_{i=1}^{HW} \sum_{k=1}^{n_c} \text{KL}(\mathbf{p}_{\text{geo}}^{s} \| \mathbf{p}_{\text{sem}}^{s}) + \text{KL}(\mathbf{p}_{\text{sem}}^{s} \| \mathbf{p}_{\text{geo}}^{s}), \\
    \mathcal{L}_{\text{SGC}}^{t} &= \sum_{i=1}^{HW} \sum_{k=1}^{n_c} \text{KL}(\mathbf{p}_{\text{geo}}^{t} \| \mathbf{p}_{\text{sem}}^{t}) + \text{KL}(\mathbf{p}_{\text{sem}}^{t} \| \mathbf{p}_{\text{geo}}^{t}). \\
    \end{aligned}
    \label{eq:eq12}
\end{equation}
Thus, the total loss of the geometry-aware adaptation stage is:
\begin{equation}
    \begin{aligned}
    \mathcal{L}_{\text{geometry-aware}} = \mathcal{L}^{s}_{\text{seg}} + \mathcal{L}^{t}_{\text{seg}} + \alpha \mathcal{L}_{\text{SGC}}^s + \beta \mathcal{L}_{\text{SGC}}^t,
    \end{aligned}
    \label{eq:eq13}
\end{equation}
where $\alpha$ and $\beta$ control the weights of the SGC loss on each domain.

\noindent\textbf{Inference.}
The Geometry-Aware Adaptation stage incorporates the complementary geometric information from 3D point cloud to strengthen final segmentation predictions.
In inference, we maintain the 2D semantic-aware predictions $\mathbf{p}^t_{\text{sem}}$ as the final prediction and discard the 3D adaptive part.
Thus, the 3D geometry-aware predictions $\mathbf{p}^t_{\text{geo}}$ is only applied to optimize $\mathbf{p}^t_{\text{sem}}$ in the Geometry-Aware Adaptation.

\section{Experiments}
\begin{table*}[t]
    \centering
    \caption{Quantitative results on GTA5 $\to$ Cityscapes in the metrics of mIoU and s-mIoU. mIoU and s-mIoU are the scores on 19 categories and 14 small-object categories, respectively. The small-object categories are highlighted with $\dagger$. The best score is highlighted \textbf{bold} and the second best one is marked {\underline{underline}} for each column.}\label{tab:tab1}
    {\fontfamily{cmr}
    \resizebox{\textwidth}{!}{
      \begin{tabular}{l|ccccccccccccccccccc|cc|cc}
      \toprule
       Method & \rotatebox{90}{road} & \rotatebox{90}{sideway}  & \rotatebox{90}{building} & {\rotatebox{90}{wall$^\dagger$}} & \rotatebox{90}{fence$^\dagger$} & \rotatebox{90}{pole$^\dagger$} & \rotatebox{90}{light$^\dagger$} & \rotatebox{90}{sign$^\dagger$} & \rotatebox{90}{vege.} & \rotatebox{90}{terrace$^\dagger$} & \rotatebox{90}{sky} & \rotatebox{90}{person$^\dagger$} & \rotatebox{90}{rider$^\dagger$} & \rotatebox{90}{car$^\dagger$} & \rotatebox{90}{truck$^\dagger$} & \rotatebox{90}{bus$^\dagger$} & \rotatebox{90}{train$^\dagger$} & \rotatebox{90}{motor$^\dagger$} & \rotatebox{90}{bike$^\dagger$} & mIoU  & gain & s-mIoU & gain \\
      \midrule
      \midrule
      Source Only & 75.8  & 16.8  & 77.2  & 12.5  & 21.0  & 25.5  & 30.1 & 20.1  & 81.3  & 24.6  & 70.3  & 53.8  & 26.4 & 49.9  & 17.2  & 25.9  & 6.5   & 25.3 & 36.0 & 36.6 &  +0.0 & 26.8 & +0.0\\
      Seg-Uncertainty~\cite{Zheng2021Uncertainty} & 90.4  & 31.2  & 85.1  & 36.9  & 25.6  & 37.5  & 48.8  & 48.5  & 85.3  & 34.8  & 81.1  & 64.4  & 36.8 & 86.3  & 34.9  & 52.2  & 1.7 & 29.0 & 44.6 & 50.3 & +13.7 & 41.6 & +14.8\\
      SAC~\cite{Araslanov2021} & 90.4 & 53.9 & 86.6 & 42.4 & 27.3 & 45.1 & 48.5 & 42.7 & 87.4 & 40.1 & 86.1 & 67.5 & 29.7 & 88.5 & 49.1 & 54.6 & 9.8 & 26.6 & 45.3 & 53.8 & +17.2 & 44.1 & +17.3\\
      Coarse-to-Fine~\cite{Ma2021} & 92.5 & 58.3 & 86.5 & 27.4 & 28.8 & 38.1 & 46.7 & 42.5 & 85.4 & 38.4 & 91.8 & 66.4 & 37.0 & 87.8 & 40.7 & 52.4 & 44.6 & 41.7 & 59.0 & 56.1 & +19.5 & 46.5 & +19.7 \\
      CorDA~\cite{Wang2021CorDA} & 94.7 & 63.1 & 87.6 & 30.7 & 40.6 & 40.2 & 47.8 & 51.6 & 87.6 & 47.0 & 89.7 & 66.7 & 35.9 & 90.2 & 48.9 & 57.5 & 0.0 & 39.8 & 56.0 & 56.6 & +20.0 & 46.6 & +19.8 \\
      ProDA~\cite{Zhang2021ProDA} & 87.8  & 56.0  & 79.7  & 46.3 & 44.8 & 45.6 & 53.5 & 53.5 & 88.6 & 45.2  & 82.1 & 70.7 & 39.2 & 88.8 & 45.5 & 59.4 & 1.0 & 48.9 & 56.4 & 57.5 & +20.9 & 49.9 & +23.1 \\
      DAFormer~\cite{Hoyer2022DAFormer} & 95.7 & 70.2 & 89.4 & 53.5 & 48.1 & 49.6 & 55.8 & 59.4 & 89.9 & 47.9 & 92.5 & 72.2 & 44.7 & 92.3 & 74.5 & 78.2 & 65.1 & 55.9 & 61.8 & 68.3 & +31.7 & 61.4 & +34.6 \\
      HRDA~\cite{Hoyer2022HRDA} & \underline{96.4} & \underline{74.4} & \underline{91.0} & \underline{61.6} & \underline{51.5} & \underline{57.1} & \underline{63.9} & \textbf{69.3} & \underline{91.3} & \underline{48.4} & \underline{94.2} & \underline{79.0} & \underline{52.9} & \underline{93.9} & \textbf{84.1} & \underline{85.7} & \textbf{75.9} & \underline{63.9} & \underline{67.5} & \underline{73.8} & \underline{+37.2} & \underline{68.2} & \underline{+41.4} \\
      \hline
      GANDA & \textbf{96.5} & \textbf{74.8} & \textbf{91.4} & \textbf{61.7} & \textbf{57.3} & \textbf{59.2} & \textbf{65.4} & \underline{68.8} & \textbf{91.5} & \textbf{49.9} & \textbf{94.7} & \textbf{79.6} & \textbf{54.8} & \textbf{94.1} & \underline{81.3} & \textbf{86.8} & \underline{74.6} & \textbf{64.8} & \textbf{68.2} & \textbf{74.5} & \textbf{{+37.9}} & \textbf{69.0} & \textbf{+42.2} \\
      \bottomrule
      \end{tabular}
    }
    }
  \end{table*}
  \begin{table*}[t]
      \centering
      \caption{Quantitative results on SYNTHIA $\to$ Cityscapes in the metrics of mIoU and s-mIoU. mIoU, mIoU* and s-mIoU are the scores on 16 categories, 13 categories and 11 small-object categories, respectively. The small-object categories are highlighted with $\dagger$. The best score is highlighted \textbf{bold} and the second best one is marked {\underline{underline}} for each column.}\label{tab:tab2}
      \resizebox{\textwidth}{!}{
        \begin{tabular}{l|cccccccccccccccc|cc|cc|cc}
        \toprule
        Method & \rotatebox{90}{road} & \rotatebox{90}{sideway} & \rotatebox{90}{building} & \rotatebox{90}{wall*$^\dagger$} & \rotatebox{90}{fence*$^\dagger$} & \rotatebox{90}{pole*$^\dagger$} & \rotatebox{90}{light$^\dagger$} & \rotatebox{90}{sign$^\dagger$} & \rotatebox{90}{vege.} & \rotatebox{90}{sky} & \rotatebox{90}{person$^\dagger$} & \rotatebox{90}{rider$^\dagger$} & \rotatebox{90}{car$^\dagger$} & \rotatebox{90}{bus$^\dagger$} & \rotatebox{90}{motor$^\dagger$} & \rotatebox{90}{bike$^\dagger$} & mIoU & gain & mIoU* & gain* & s-mIoU & gain\\
        \midrule
        \midrule
        Source Only & 64.3 & 21.3 & 73.1 & 2.4 & 1.1 & 31.4 & 7.0 & 27.7 & 63.1 & 67.6 & 42.2 & 19.9 & 73.1 & 15.3 & 10.5 & 38.9 & 34.9 & +0.0 & 40.3 & +0.0 & 24.5 & +0.0\\
        GIO-Ada~\cite{Chen2019} & 78.3 & 29.2 & 76.9 & 11.4 & 0.3 & 26.5 & 10.8 & 17.2 & 81.7 & 81.9 & 45.8 & 15.4 & 68.0 & 15.9 & 7.5 & 30.4 & 37.3 & +2.4 & 43.0 & +2.7 & 22.7 & -1.8 \\
        DADA~\cite{Vu2019DADA} & 89.2 & 44.8 & 81.4 & 6.8 & 0.3 & 26.2 & 8.6 & 11.1 & 81.8 & 84.0 & 54.7 & 19.3 & 79.7 & 40.7 & 14.0 & 38.8 & 42.6 & +7.7 & 49.8 & +9.5 & 27.3 & +2.8 \\
        Seg-Uncertainty~\cite{Zheng2021Uncertainty} & 87.6 & 41.9 & 83.1 & 14.7 & 1.7 & 36.2 & 31.3 & 19.9 & 81.6 & 80.6 & 63.0 & 21.8 & 86.2 & 40.7 & 23.6 & 53.1 & 47.9 & +13.0 & 54.9 & +14.6 & 35.7 & +11.2\\
        Coarse-to-Fine~\cite{Ma2021} & 75.7 & 30.0 & 81.9 & 11.5 & 2.5 & 35.3 & 18.0 & 32.7 & 86.2 & 90.1 & 65.1 & 33.2 & 83.3 & 36.5 &35.3 & 54.3 & 48.2 & +13.3 & 55.5 & +15.2 & 37.1 & +12.6 \\
        SAC~\cite{Araslanov2021} & \underline{89.3} & 47.2 & 85.5 & 26.5 & 1.3 & 43.0 & 45.5 & 32.0 & \underline{87.1} & 89.3 & 63.6 & 25.4 & 86.9 & 35.6 & 30.4 & 53.0 & 51.6 & +16.7 & 59.3 & +19.0 & 40.3 & +15.8 \\
        CorDA~\cite{Wang2021CorDA} & \textbf{93.3} & \textbf{61.6} & 85.3 & 19.6 & 5.1 & 37.8 & 36.6 & 42.8 & 84.9 & 90.4 & 69.7 & 41.8 & 85.6 & 38.4 & 32.6 & 53.9 & 55.0 & +20.1 & 62.8 & +22.5 & 42.2 & +17.7 \\
        ProDA~\cite{Zhang2021ProDA} & 87.8 & 45.7 & 84.6 & 37.1 & 0.6 & 44.0 & 54.6 & 37.0 & \textbf{88.1} & 84.4 & 74.2 & 24.3 & 88.2 & 51.1 & 40.5 & 45.6 & 55.5 & +20.6 & 62.0 & +21.7 & 45.2 & +20.7\\
        DAFormer~\cite{Hoyer2022DAFormer} & 84.5 & 40.7 & 88.4 & 41.5 & \underline{6.5} & 50.0 & 55.0 & 54.6 & 86.0 & 89.8 & 73.2 & 48.2 & 87.2 & 53.2 & 53.9 & 61.7 & 60.9 & +26.0 & 67.4 & +27.1 & 53.2 & +28.7 \\
        HRDA~\cite{Hoyer2022HRDA} & 85.2 & 47.7 & \underline{88.8} & \underline{49.5} & 4.8 & \underline{57.2} & \underline{65.7} & \textbf{60.9} & 85.3 & \underline{92.9} & \underline{79.4} & \underline{52.8} & \underline{89.0} & \underline{64.7} & \textbf{{63.9}} & \underline{64.9} & \underline{65.8} & \underline{+30.9} & \underline{72.4} & \underline{+32.1} & \underline{59.3} & \underline{+34.8}\\
        \hline
        GANDA & 89.1 & \underline{50.6} & \textbf{89.7} & \textbf{51.4} & \textbf{6.7} & \textbf{59.4} & \textbf{66.8} & \underline{57.7} & 86.7 & \textbf{93.8} & \textbf{80.6} & \textbf{56.9} & \textbf{90.7} & \textbf{64.8} & \underline{62.6} & \textbf{65.0} & \textbf{67.0} & \textbf{+32.1} & \textbf{73.5} & \textbf{+33.2} & \textbf{60.2} & \textbf{+35.7} \\
        \bottomrule
        \end{tabular}
      }
    \end{table*}
Our proposed GANDA is evaluated on the benchmarks SYNTHIA$\to$Cityscapes and GTA5$\to$Cityscapes.
\subsection{Implementation}
\noindent\textbf{Datasets.} In our experiments, we adopt the GTA5 and SYNTHIA datasets as the source synthetic datasets, respectively, and exploit the Cityscapes dataset as the target real scene dataset.
The GTA5 dataset~\cite{StephanR2016GTA} has 24,966 synthetic photo-realistic scenes with a resolution of 1914 $\times$ 1052.
SYNTHIA~\cite{Ros2016SYN} is a large-scale synthetic urban image dataset, containing 9,400 synthetic urban images with the resolution of 1280 $\times$ 760 in model training.
The Cityscapes dataset~\cite{Cordts2016City} is a realistic dataset of the 2048 $\times$ 1024 street scenes from 50 cities, where 2,975 images for training and 500 ones for validation.

\noindent\textbf{Setups.} Two synthetic datasets both have pixel-level segmentation annotations, and these labels are utilized to provide supervision on the source data.
Besides, the GTA5 dataset shares 19 semantic categories, and the SYNTHIA dataset shares 16 common classes with the Cityscapes dataset, respectively.
%
% Following the previous works~\cite{Zhang2021ProDA}, we use the images in the synthetic dataset and the train split of the Cityscapes in training, and we apply the validation split of the Cityscapes for the final evaluation.
%
We compute the Intersection-over-Union (IoU) on each class and derive the mean IoU (mIoU) for the evaluation metric.
To evaluate the performances on the small-object categories, we further derive the small-object mean IoU (\textbf{s-mIoU}) that computes the average IoU on the small-object categories, and demonstrate the segmentation improvements on the small-object categories.

\noindent\textbf{Calibrated Camera Intrinsics.} The intrinsic parameters for the images are provided in the datasets or can be estimated using the software like RenderDoc.
Thus, in the transformation from RGBD to point cloud, we specifically consider and utilize the calibrated camera intrinsics for each domain.
%
% For example, an image in Cityscapes dataset provides the camera parameters: the pixel focal length $f_x$ = 2268.36, $f_y$ = 2225.54 and the principal point $u_0$ = 1048.64, $v_0$ = 519.28.
%
% These parameters are used to obtain the intrinsic matrix.

\noindent\textbf{Implementation Details.} Our model is implemented in PyTorch.
Following the previous works~\cite{Zhang2021ProDA,Hoyer2022HRDA}, we utilize the DeepLabv2~\cite{Chen2018DeepLabv2} and SegFormer~\cite{Xie2021SegFormer} as the segmentation architectures.
%
% In the depth-aware adversarial adaptation stage, we follow the architecture of the depth decoder in the Monodepth2~\cite{Godard2019} and adapt it with the input image size as 896 $\times$ 512 in ~\cite{Zhang2021ProDA} or 512 $\times$ 512 in ~\cite{Hoyer2022HRDA}.
%
% The SGD optimizer is utilized with a learning rate of 0.0002, a decay of 0.0002 for every epoch.
%
% The total number of training iterations is 50,000, and the batch size is 2.
% %
On GTA5 $\to$ Cityscapes, since the GTA5 dataset has no depth maps, we obtain the depths via training the NeWCRFs~\cite{Yuan2022NeWCRFs} on the GTA5-based dataset DeepMVS~\cite{Huang2018DeepMVS}.
On SYNTHIA $\to$ Cityscapes, we utilize the provided depth maps from Synthia-Rand-Cityscapes.
% %
% In the geometry-aware adaptation stage, we discard the parameters in the depth decoder and preserve the parameters of the DeepLabv2 or the SegFormer.
%
\begin{figure}[t]
  \centering
  \includegraphics[width=1.00\columnwidth]{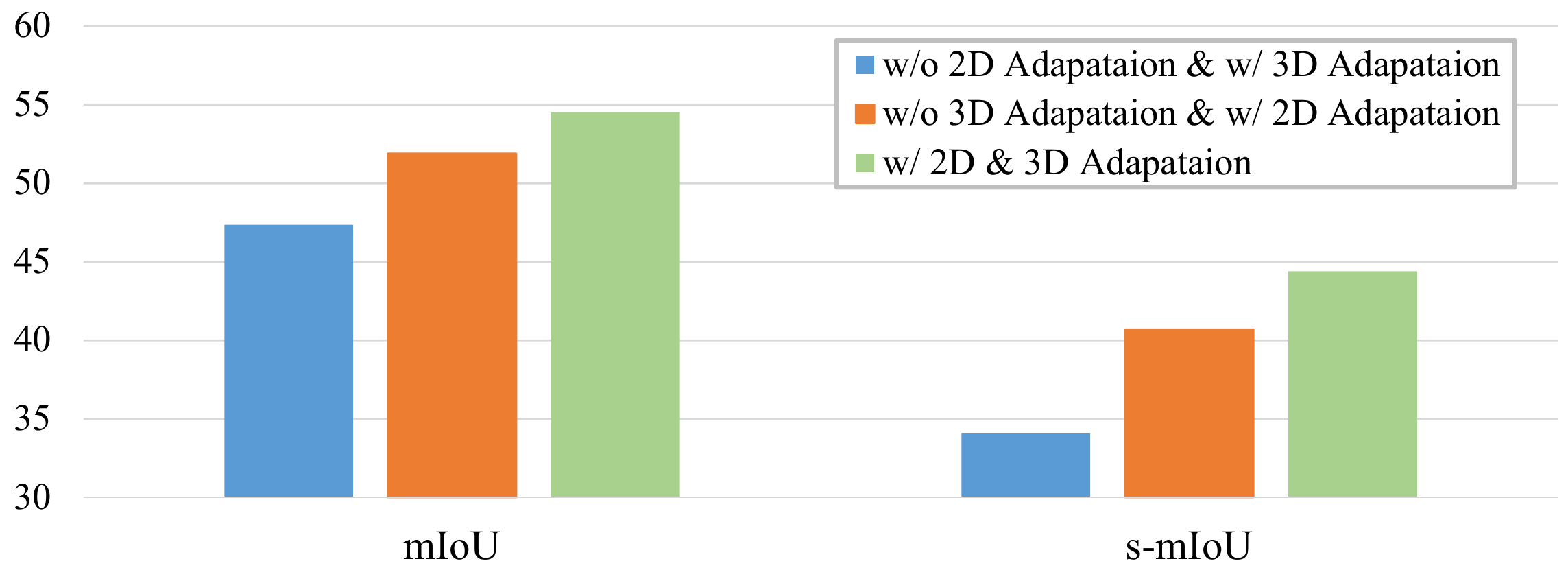}
  \caption{Comparisons with and without the 2D and 3D segmentation on GANDA+ProDA on the SYNTHIA$\to$Cityscapes task in the Geometry-Aware Adaptation stage in the metrics of mIoU and s-mIoU.}
  \label{fig:fig5}
\end{figure}
%
% The radius parameters of 4 set abstraction (SA) layers in the PointNet++~\cite{Qi2017PointNet} are set as 0.003, 0.006, 0.012, and 0.024 in SYNTHIA$\to$Cityscapes and set as 0.0008, 0.0016, 0.0032, and 0.0064 in GTA5$\to$Cityscapes.
% %
% Unlike the default settings in PointNet++, we modify the dimension of point-aware features from 128 to 256.
% %
% At the stage of geometry-aware adaptation, the SGD optimizer is applied with a learning rate of 0.0001, a decay of 0.0002, and a momentum of 0.9 for every epoch.
% %
% The total number of training epochs is 150, and the batch size is 4, including the images and point clouds.
%
% The weight of the smoothness loss is set as $\lambda = 0.01$, the weight of the adversarial loss is set as $\mu = 0.001$ and the weights of the SGC loss are set as $\alpha = 0.5$ and $\beta = 0.25$, respectively.
%
We conduct the experiments on the Tesla V100 GPUs.
\subsection{Comparisons with State-Of-The-Arts}
\begin{table*}[t]
      \centering
      \caption{Effectiveness of the Geometry-Aware Adaptation in multiple UDA methods on SYNTHIA $\to$ Cityscapes in the metrics of mIoU, mIoU* and s-mIoU. mIoU, mIoU* and s-mIoU are the scores on 16 categories, 13 categories and 11 small-object categories, respectively.  The symbol ``$\star$" indicates the distillation operation is applied.}\label{tab:tab3}
      \resizebox{\textwidth}{!}{
        \begin{tabular}{l|cccccccccccccccc|c|c|c}
        \toprule
        Method & \rotatebox{90}{road} & \rotatebox{90}{sideway} & \rotatebox{90}{building} & \rotatebox{90}{wall*$^\dagger$} & \rotatebox{90}{fence*$^\dagger$} & \rotatebox{90}{pole*$^\dagger$} & \rotatebox{90}{light$^\dagger$} & \rotatebox{90}{sign$^\dagger$} & \rotatebox{90}{vege.} & \rotatebox{90}{sky} & \rotatebox{90}{person$^\dagger$} & \rotatebox{90}{rider$^\dagger$} & \rotatebox{90}{car$^\dagger$} & \rotatebox{90}{bus$^\dagger$} & \rotatebox{90}{motor$^\dagger$} & \rotatebox{90}{bike$^\dagger$} & mIoU & mIoU* & s-mIoU \\
        \midrule
        \midrule
        \multicolumn{20}{l}{\emph{Self-training Method~~~~~~~~~~~~~~~~~~~~~~~~~~~~~~~~~~~~~~~~~~~~Architecture: DeepLabv2~~~~~~~~~~~~~~~~~~~~Backbone:ResNet-101}}\\
        \midrule
        ProDA~\cite{Zhang2021ProDA} & 87.0 & 44.0 & 83.2 & 26.9 & 1.0 & 42.0 & 45.8 & 34.2 & 86.7 & 81.3 & 68.4 & 22.1 & 87.7 & 50.0 & 31.4 & 38.6 & 51.9 & 58.5 & 40.7 \\
        GANDA + ProDA & 87.6 & 44.2 & 84.0 & 31.0 & 1.0 & 42.0 & 50.5 & 38.5 & 87.1 & 81.2 & 68.5 & 27.2 & 88.7 & 50.2 & 44.4 & 46.6 & 54.5 & 61.4 & 44.4\\
        Gain & \textbf{$\uparrow$0.6} & \textbf{$\uparrow$0.2} & \textbf{$\uparrow$0.8} & \textbf{$\uparrow$4.1} & \textbf{$\uparrow$0.0} & \textbf{$\uparrow$0.0} & \textbf{$\uparrow$4.7} & \textbf{$\uparrow$4.3} & \textbf{$\uparrow$0.4} & {$\downarrow$0.1} & \textbf{$\uparrow$0.1} & \textbf{$\uparrow$5.1} & \textbf{$\uparrow$1.0} & \textbf{$\uparrow$0.2} & \textbf{$\uparrow$13.0} & \textbf{$\uparrow$8.0} & \textbf{$\uparrow$2.6} & \textbf{$\uparrow$2.9} & \textbf{$\uparrow$3.7}\\
        \midrule
        \multicolumn{20}{l}{\emph{Self-training Method with Distillation~~~~~~~~~~~~~~~~~Architecture: DeepLabv2~~~~~~~~~~~~~~~~~~~~Backbone:ResNet-101}}\\
        \midrule
        ProDA$^\star$~\cite{Zhang2021ProDA} & 87.8 & 45.7 & 84.6 & 37.1 & 0.6 & 44.0 & 54.6 & 37.0 & 88.1 & 84.4 & 74.2 & 24.3 & 88.2 & 51.1 & 40.5 & 45.6 & 55.5 & 62.0 & 45.2 \\
        GANDA + ProDA$^\star$ & 87.6 & 44.8 & 85.6 & 36.5 & 0.0 & 45.8 & 50.3 & 40.0 & 88.2 & 86.4 & 71.1 & 30.1 & 90.8 & 57.4 & 40.6 & 52.9 & 56.8 & 63.5 & 46.9 \\
        Gain & {$\downarrow$0.2} & {$\downarrow$0.9} & \textbf{$\uparrow$1.0} & {$\downarrow$0.6} & {$\downarrow$0.6} & \textbf{$\uparrow$1.8} & {$\downarrow$4.3} & {$\downarrow$3.0} & \textbf{$\uparrow$0.1} & \textbf{$\uparrow$2.0} & {$\downarrow$3.1} & \textbf{$\uparrow$5.8} & \textbf{$\uparrow$2.6} & \textbf{$\uparrow$6.3} & \textbf{$\uparrow$0.1} & \textbf{$\uparrow$7.3} & \textbf{$\uparrow$1.3} & \textbf{$\uparrow$1.5} & \textbf{$\uparrow$1.7} \\
        \midrule
        \multicolumn{20}{l}{\emph{Depth-based Multi-task Learning Method~~~~~~~~~~Architecture: DeepLabv2~~~~~~~~~~~~~~~~~~~~Backbone:ResNet-101}}\\
        \midrule
        CorDA~\cite{Wang2021CorDA} & 93.3 & 61.6 & 85.3 & 19.6 & 5.1 & 37.8 & 36.6 & 42.8 & 84.9 & 90.4 & 69.7 & 41.8 & 85.6 & 38.4 & 32.6 & 53.9 & 55.0 & 62.8 & 42.2 \\
        GANDA + CorDA & 87.1 & 45.8 & 86.1 & 28.9 & 4.8 & 37.1 & 40.6 & 45.0 & 87.0 & 87.9 & 69.1 & 39.8 & 89.9 & 59.8 & 33.8 & 57.2 & 56.3 & 63.8 & 46.0 \\
        Gain & {$\downarrow$6.2} & {$\downarrow$15.8} & \textbf{$\uparrow$0.8} & \textbf{$\uparrow$9.3} & {$\downarrow$0.3} & {$\downarrow$0.7} & \textbf{$\uparrow$4.0} & \textbf{$\uparrow$2.2} & \textbf{$\uparrow$2.1} & {$\downarrow$2.5} & {$\downarrow$0.6} & {$\downarrow$2.0} & \textbf{$\uparrow$4.3} & \textbf{$\uparrow$21.4} & \textbf{$\uparrow$1.2} & \textbf{$\uparrow$3.3} & \textbf{$\uparrow$1.3} & \textbf{$\uparrow$1.0} & \textbf{$\uparrow$3.8} \\
        \midrule
        \multicolumn{20}{l}{\emph{Self-training Method~~~~~~~~~~~~~~~~~~~~~~~~~~~~~~~~~~~~~~~~~~~~Architecture: SegFormer~~~~~~~~~~~~~~~~~~~~~Backbone:MiT-B5}}\\
        \midrule
        HRDA~\cite{Hoyer2022HRDA} & 85.2 & 47.7 & 88.8 & 49.5 & 4.8 & 57.2 & 65.7 & 60.9 & 85.3 & 92.9 & 79.4 & 52.8 & 89.0 & 64.7 & 63.9 & 64.9 & 65.8 & 72.4 & 59.3 \\
        GANDA + HRDA & 89.1 & 50.6 & 89.7 & 51.4 & 6.7 & 59.4 & 66.8 & 57.7 & 86.7 & 93.8 & 80.6 & 56.9 & 90.7 & 64.8 & 62.6 & 65.0 & 67.0 & 73.5 & 60.2 \\
        Gain & \textbf{$\uparrow$3.9} & \textbf{$\uparrow$2.9} & \textbf{$\uparrow$0.9} & \textbf{$\uparrow$1.9} & \textbf{$\uparrow$1.9} & \textbf{$\uparrow$2.2} & \textbf{$\uparrow$1.1} & {$\downarrow$3.2} & \textbf{$\uparrow$1.4} & \textbf{$\uparrow$1.1} & \textbf{$\uparrow$1.2} & \textbf{$\uparrow$4.1} & \textbf{$\uparrow$1.7} & \textbf{$\uparrow$0.1} & {$\downarrow$1.3} & \textbf{$\uparrow$0.1} & \textbf{$\uparrow$1.2} & \textbf{$\uparrow$1.1} & \textbf{$\uparrow$0.9} \\
        \bottomrule
        \end{tabular}
      }
    \end{table*}
\begin{table}
\footnotesize
  \caption{Ablation studies of each component on the GANDA+ProDA on the SYNTHIA $\to$ Cityscapes task. AA and GAA are the Adversarial Adaptation and Geometric-Aware Adaptation. ST, GA, WCE, and SGC denote Self-Training Only, Geometry Adaptation module, Weighted CE loss, and Semantic-Geometric Consistency loss.}\label{tab:tab4}
  \centering
\begin{tabular}{c|cccc|cc}
\toprule
 \multicolumn{5}{c|}{Stage \& Components}   & \multicolumn{1}{c}{mIoU} & \multicolumn{1}{c}{Gain} \\
\midrule
\midrule
    \multicolumn{5}{c|}{Source Only}  & 34.9  & +0.0 \\
\midrule
    \multirow{1}[4]{*}{AA} & \multicolumn{4}{c|}{Image Only}  & 40.9  & +6.0 \\
     & \multicolumn{4}{c|}{Depth-Aware}  & 41.2  & +6.3 \\
\midrule
\multirow{6}[4]{*}{GAA} & \multicolumn{1}{c}{ST} & \multicolumn{1}{c}{GA} & \multicolumn{1}{c}{WCE} & \multicolumn{1}{c|}{SGC} & \multicolumn{1}{c}{mIoU} & \multicolumn{1}{c}{Gain} \\
\cline{2-7}
& \cmark &        &       &         & 51.9 & +17.0 \\
& \cmark &    &       & \cmark  & 52.3 & +17.4 \\
& \cmark & \cmark &       &         & 52.9  & +18.0 \\
& \cmark &    & \cmark &        & 53.1  & +18.2 \\
& \cmark & \cmark & \cmark & \cmark  & 54.5  & +19.6 \\
% \midrule
%     \multicolumn{5}{c|}{Distillation}  & 58.0  & +21.4 \\
\bottomrule
  \end{tabular}
\end{table}
\begin{table}[h]
\footnotesize
\caption{Computation comparisons in HRDA with and without Geometry-Aware Adaptation in the training and inference phases with the metrics of FLOPs (G) and Parameters (Params) (M) on the 512$\times$512 input image and the extra 40000 input points in training.}\label{tab:tab5}
\centering
\resizebox{0.45\textwidth}{!}{
\begin{tabular}{l|cc|cc} 
\toprule
\multirow{2}[1]{*}{Model} & \multicolumn{2}{c|}{Training} & \multicolumn{2}{c}{Inference}\\
& FLOPs (G) & Params (M) & FLOPs (G) & Params (M)\\
\midrule
\midrule
HRDA & 146.9 & 85.7 & \multirow{2}[1]{*}{146.9} & \multirow{2}[1]{*}{85.7} \\
GANDA + HRDA & 150.4 & 86.7 \\
\bottomrule
\end{tabular}
}
\end{table}
The evaluations of our GANDA are conducted via the comparisons of several state-of-the-art methods on the two scenarios: GTA5$\to$Cityscapes in Table~\ref{tab:tab1} and SYNTHIA $\to$Cityscapes in Table~\ref{tab:tab2}.
These methods are mainly divided into two categories.
a) \textbf{Self-training Methods}: Seg-Uncertainty~\cite{Zheng2021Uncertainty}, SAC~\cite{Araslanov2021}, Coarse-to-Fine~\cite{Ma2021}, ProDA~\cite{Zhang2021ProDA}, DAFormer~\cite{Hoyer2022DAFormer}, and HRDA~\cite{Hoyer2022HRDA}.
b) \textbf{Depth-based Multi-task Learning Methods}: GIO-Ada~\cite{Chen2019}, DADA~\cite{Vu2019DADA}, and CorDA~\cite{Wang2021CorDA}.
On the GTA5 $\to$ Cityscapes task in Table~\ref{tab:tab1}, our proposed GANDA outperforms of the state-of-the-art HRDA by 0.7\% mIoU and 0.8\% s-mIoU.
On the more challenging SYNTHIA $\to$ Cityscapes task in Table~\ref{tab:tab2}, our proposed GANDA achieves the highest mIoU score as 67.0\%, 73.5\%, and 60.2\% on each metric, respectively.
%
% Though HRDA is effective in the segmentation of small objects, our GANDA outperforms HRDA by nearly 1\% IoU on each small-object class.
%
The performances on the two benchmarks well illustrate the capability of GANDA in shrinking the domain gaps and leveraging the geometric information to perceive the shapes and spatial positions of small objects.
\subsection{Effectiveness of Geometry-Aware Adaptation}
Since our GANDA is a plug-and-play UDA segmentation framework that can be applied to any existed UDA method, we perform the experiments on the multiple UDA methods with and without the Geometry-Aware Adaptation in Table~\ref{tab:tab3}, including self-training ProDA~\cite{Zhang2021ProDA}, depth-based CorDA~\cite{Wang2021CorDA}, and self-training HRDA~\cite{Hoyer2022HRDA}.
The gains by adding GANDA are evident in each method in terms of all the metrics, especially the gains on s-mIoU demonstrate the efficacy of differentiating the small object classes. 
\subsection{Ablation Study}
The proposed components of our GANDA are evaluated based on their contributions to the model performances in Table~\ref{tab:tab4}, including Depth-Aware Adversarial Adaptation, Geometry Adaptation module (GA), Weighted CE loss (WCE), and Semantic-Geometric Consistency loss (SGC).
According to Table~\ref{tab:tab4}, at the stage of Adversarial Adaptation (AA), the introduction of depth estimation can gain the performances of 43.8\% mIoU.
At the stage of Geometry-Aware Adaptation (GAA), the model using only the Self-Training (ST) techniques in ProDA~\cite{Zhang2021ProDA} has a gain of 16.9\% mIoU compared to the source only model.
After adding the GA, WCE, and SGC, the model performances are gradually increasing and achieving the gains from 17.6\% to 18.5\%.
Besides, we perform the comparisons with and without the 2D and 3D segmentation on GANDA+ProDA on the SYNTHIA$\to$Cityscapes task in the Geometry-Aware Adaptation stage in Figure~\ref{fig:fig5}.
It proves that 2D semantic segmentation is necessary and the 3D geometric part is complementary to classifying the small-object classes.
\subsection{Computation Costs in Training and Inference}
Table~\ref{tab:tab5} presents the computation costs in training and inference in HRDA with and without Geometry-Aware Adaptation in terms of FLOPs and parameters.
Inclusion of the GANDA does not lead to the large computation costs, since it is only applied in training and discarded in inference.
Despite the increased FLOPs and model parameters in training, the extra computation cost is acceptable.
\section{Conclusions}
In this paper, we address the task of synthetic-to-real domain adaptive semantic segmentation by proposing a two-stage geometry-aware training framework.
Beyond the previous depth-based UDA methods and based only on source depth information, GANDA gains a deep insight into the 3D geometric space and extracts the domain-invariant information after the structure-texture and coordinate-color disentanglements.
Besides, without any structure or model dependency, GANDA is plug-and-play and boosts the performances of other existing UDA methods, especially effective in the segmentation of small-object categories.
Comprehensive experimental results and analysis demonstrate the effectiveness of our method, where GANDA outperforms other methods on two prevalent synthetic-to-real tasks.
We hope this work will prompt more researchers to explore the potential of geometric information on the UDA tasks.
\section{Acknowledgments}
This work was supported in part by JCYJ20220530143600001, by the Basic Research Project No. HZQB-KCZYZ-2021067 of Hetao Shenzhen HK S\&T Cooperation Zone, by the National Key R\&D Program of China with grant No.2018YFB1800800, by SGDX20211123112401002, by Shenzhen Outstanding Talents Training Fund, by Guangdong Research Project No. 2017ZT07X152 and No. 2019CX01X104, by the Guangdong Provincial Key Laboratory of Future Networks of Intelligence (Grant No. 2022B1212010001), by the NSFC 61931024\&8192 2046, by NSFC-Youth 62106154, by zelixir biotechnology company Fund, by Tencent Open Fund, and by ITSO at CUHKSZ.
\bibliography{aaai23}

\end{document}